\documentclass[table]{article}

\usepackage{arxiv}

\usepackage{amsmath}

\usepackage[utf8]{inputenc} 
\usepackage[T1]{fontenc}    
\usepackage{hyperref}       
\usepackage{url}            
\usepackage{booktabs}       
\usepackage{amsfonts}       
\usepackage{nicefrac}       
\usepackage{microtype}      
\usepackage{cleveref}       
\usepackage{graphicx}
\usepackage[round]{natbib}
\usepackage{doi}

\usepackage{tikz}
\usepackage{tcolorbox}
\usepackage{mathtools}
\usepackage{amsthm}
\usepackage{bbm}
\usepackage{bm}
\usepackage{tabularx}
\usepackage{threeparttable}
\usepackage{makecell}
\usepackage{multirow}
\usepackage{booktabs}
\usepackage{subcaption}
\usepackage{flushend}
\usepackage{dirtytalk}
\usepackage{xcolor}

\definecolor{fhgreen}{RGB}{0,147,116}
\definecolor{orange}{RGB}{	255, 127, 0}
\definecolor{unibonnblue}{HTML}{0a55a1}

\title{Controlled Randomness Improves the Performance of Transformer Models}

\date{}

\usepackage{authblk}

\author[1,2]{%
	Tobias Deu{\ss}er\thanks{\texttt{tdeusser@uni-bonn.de}, ORCID-ID: 0000-0003-4685-0847}%
}
\author[1,3]{%
	Cong Zhao
}
\author[3]{%
	Wolfgang Krämer
}
\author[1,2]{%
	David Leonhard
}
\author[1,2]{%
	\\Christian Bauckhage
}
\author[1,2]{%
	Rafet Sifa
}
\affil[1]{University of Bonn, Bonn, Germany}
\affil[2]{Fraunhofer IAIS, Sankt Augustin, Germany}
\affil[3]{Deutsche Telekom, Bonn, Germany}

\renewcommand{\shorttitle}{Controlled Randomness Improves the Performance of Transformer Models}

\hypersetup{
pdftitle={Controlled Randomness Improves the Performance of Transformer Models},
pdfsubject={cs.CL, cs.AI},
pdfauthor={Tobias Deu{\ss}er, Cong Zhao, Wolfgang Krämer, David Leonhard, Christian Bauckhage, Rafet Sifa},
pdfkeywords={Natural Language Processing, Regularization, Transformer, Machine Learning},
}

\newcommand\copyrighttext{%
  \footnotesize \textcopyright 2023 IEEE. Personal use of this material is permitted. Permission from IEEE must be obtained for all other uses, in any current or future media, including reprinting/republishing this material for advertising or promotional purposes, creating new collective works, for resale or redistribution to servers or lists, or reuse of any copyrighted component of this work in other works.
}
\newcommand\copyrightnotice{%
\begin{tikzpicture}[remember picture,overlay]
\node[anchor=south,yshift=10pt] at (current page.south) {\fbox{\parbox{\dimexpr\textwidth-\fboxsep-\fboxrule\relax}{\copyrighttext}}};
\end{tikzpicture}%
}

\begin{document}
\maketitle
\addtocounter{footnote}{-1}
\begin{abstract}
During the pre-training step of natural language models, the main objective is to learn a general representation of the pre-training dataset, usually requiring large amounts of textual data to capture the complexity and diversity of natural language. Contrasting this, in most cases, the size of the data available to solve the specific downstream task is often dwarfed by the aforementioned pre-training dataset, especially in domains where data is scarce. We introduce controlled randomness, i.e. noise, into the training process to improve fine-tuning language models and explore the performance of targeted noise in addition to the parameters of these models. We find that adding such noise can improve the performance in our two downstream tasks of \textit{joint named entity recognition and relation extraction} and \textit{text summarization}.
\end{abstract}
\copyrightnotice

\keywords{Natural Language Processing \and Regularization \and Transformer \and Machine Learning}

\section{Introduction}

The emergence of pre-trained transformer models brought a massive breakthrough in the field of natural language processing. During pre-training, such transformer models can learn generic language representations with strong generalization capabilities by applying a self-supervised learning approach and leveraging large text corpora. These pre-trained language models can be fine-tuned in various downstream tasks without needing to train from scratch compared to traditional training methods, significantly reducing training costs while achieving excellent performance. Models like BERT \cite{devlin2018bert}, ELECTRA \cite{Clark2020ELECTRA}, or T5 \cite{t5} have achieved remarkable results on several language processing tasks and the most recent developments of even larger language models, made prominent by GPT-3  \cite{gpt3} and GPT-4 \cite{openai2023gpt4} but not limited to these two\footnote{See for example OPT \cite{zhang2022opt}, Bloom \cite{workshop2023bloom}, LLaMA \cite{touvron2023llama}, or Falcon \cite{falcon}.}, improved on these even further. These models have enabled researchers and developers to exploit existing computational linguistic knowledge more conveniently, which in turn has dramatically accelerated the development of natural language processing research and applications.

One of the keys to the success of these models is the ability to adapt to data not encountered during the pre-training phase, i.e. when a downstream task is tackled and the model fine-tuned, as shown in virtually all such applications of transformer models (e.g. \cite{chalkidis2019neural}, \cite{beltagy2020longformer}, \cite{wang2021automated}, \cite{aghajanyan2021muppet}, \cite{kpibert}, \cite{ye2022packed}, \cite{deusser2023contradiction} \cite{ramamurthy2023is}, \cite{deusser2023informed}). To avoid overfitting and instability during this process, one can apply various regularization techniques and data augmentation, which both might help stabilize fine-tuning and improve performance. 

In this work, we build upon the concept introduced in \cite{wu2022noisytune}, namely NoisyTune, which describes the process of adding noise to all the parameters of language models in order to regularize it.

We introduce such noise to more parts of the model to examine how this influences the performance of two downstream tasks, specifically \textit{joint named entity recognition and relation extraction} and \textit{text summarization}. We observe that we can enhance the F$_1$ score of the \textit{joint named entity recognition and relation extraction} model on the KPI-EDGAR dataset \cite{deusser2022kpiedgar} by \textbf{2.977} and the ROUGE--Average\footnote{See subsection~\ref{subsection:data_and_downstream_tasks} for how we define this metric.} score of the \textit{text summarization} model by \textbf{1.387} on the BillSum dataset \cite{kornilova-eidelman-2019-billsum}.

Our contribution is thus investigating how noise added to various parts can improve the performance of the considered downstream task. Therein, we find
that the model's performance is significantly improved by adding controlled noise to certain components. We further theorize that this novel approach can help improve the generalisation of large language models to small datasets or low-resource languages.

In the following, we first review related work with a focus on regularization techniques for natural language processing as well as previous studies on \textit{joint named entity recognition and relation extraction} and \textit{text summarization}. Section~\ref{section:methodology} describes our methodology, i.e., how and where we introduce noise to the model and the general model architecture of both downstream tasks. Thereafter, in Section~\ref{section:experiments}, we outline our dataset, present our experiments, and discuss the results. Section~\ref{section:conclusion} then adds concluding remarks and an outlook into conceivable future work.

\section{Related Work}
In this section, we discuss various other studies conducted on the effect of regularization techniques on machine learning models. Afterwards, we shortly introduce the most relevant advances in both of our tasks, \textit{joint named entity recognition and relation extraction} and \textit{text summarization}.

\subsection{Regularization}

Overfitting and thus regularization of machine learning models have been thoroughly studied since the emergence of the field. Early examples of regularization include the work by \cite{hoerl1970ridge}, \cite{hanson1988comparing}, \cite{mccloskey1989catastrophic}, \cite{breiman1995better}, \cite{girosi1995regularization}, and \cite{prechelt2002early}.
Using noise as a regularization technique for training machine learning models was first considered in \cite{bishop1995training}.

\cite{hinton2012improving} introduced the popular dropout method, which randomly omits a certain part of the feature detectors on each training case. Building on the idea of weight decay \cite{hanson1988comparing}, \cite{loshchilov2018decoupled} investigated the effects of decoupled weight decay regularization on the training of deep neural networks.

Applying such techniques to natural language processing has gained more importance with the ever-increasing 
size of the number of parameters of language models. \cite{merity2018regularizing} considered the
challenge of word-level language modelling and investigated strategies for regularizing Long Short Term Memory-based \cite{lstm} models. \cite{Lee2020Mixout} 
proposed to add dropout to randomly mix pre-trained parameters into the downstream model to reduce forgetting in BERT fine-tuning. Furthermore, \cite{dodge2020finetuning} 
proposed an early stopping method to filter out poor-performing random seeds. On the topic of machine translation, \cite{ott2018analyzing} analyzed uncertainty in machine translation and proposed tools to assess model calibration. \cite{jiang-etal-2020-smart} 
used smoothness-inducing regularization, which tries to effectively manage the complexity of the model to encourage models to be smooth within neighbourhoods of all the inputs.
It was theorized by by \cite{sun2019fine} that smaller learning rates during fine-tuning help the model retain prior knowledge while still adapting to the new task. In \cite{howard-ruder-2018-universal}, the authors 
proposed a gradual unfreezing strategy in which layers of the pre-trained language model are unfrozen one at a time during fine-tuning to cope with catastrophic forgetting. \cite{pan2023improving} 
introduced an extra class-aware initialization stage before fine-tuning and concluded that, in this way, self-supervised models should be easier to train to discriminate between different classes.
Finally, \cite{wu2022noisytune} described how to regularize a language model by adding noise to its weight parameters, which we will use as a further baseline for our approach. By investigating the effect of noise injection into the individual parts of the model explicitly and measuring the performance on two discrete downstream tasks, our work succeeds \cite{wu2022noisytune} significantly by adding specificity to the analysis.

\subsection{Joint Named Entity Recognition and Relation Extraction}

When it comes to the first of the evaluated downstream natural language processing tasks, there is introductory work by \cite{miwa2014modelling},  \cite{li2014incremental}, and \cite{gupta2016table} that was further confirmed by \cite{kambar2022survey}, demonstrating the advantage of joining the subtasks of named entity recognition (NER) and relation extraction (RE) together. Studies, evaluating further concepts to improve the performance of joint NER and RE (also called JNERE), were carried out by \cite{zheng2017joint}, who introduced a novel tagging scheme, \cite{bekoulis2018joint} and \cite{geng2021joint}, who treated the task as a multi-head selection problem, \cite{fu2019graphrel}, who used graph convolutional networks (GCNs), \cite{giorgi2019end} and \cite{xue2019fine}, who leveraged the pre-trained model BERT, \cite{yu2019joint}, who improved performances by introducing a new task decomposition strategy, and \cite{shang2022onerel}, who focused on the problems of cascading errors and redundant information in previous models by creating their own, OneRel, which treats joint extraction as a fine-grained triple classification
problem.

There is also insight on the application of JNERE regarding work by \cite{bhatia2019comprehend}, \cite{chen2020joint}, and \cite{jabbari2020french} for the medical, legal, and financial language domain respectively.

The work by \cite{deusser2022kpiedgar}, which was already mentioned in the introduction, also lays groundwork for the examinations undertaken in this study. However, to the best of our knowledge, the effect of deliberately incorporating noise into the training process of JNERE models specifically has not yet been
systematically
researched.

\subsection{Text Summarization}

As for the task of text summarization, there are several studies conducted on the same dataset as our experiments, that is BillSum, introduced by \cite{kornilova-eidelman-2019-billsum}. Most of these make use of or compare their proposed models to models based on BERT, such as \cite{an2021retrievalsum}, who used a model that incorporates additional knowledge into the summarization task, \cite{abdel2022performance}, who compared various BERT variants, \cite{liang2022efficient}, who proposed the Coarse-to-
Fine Facet-Aware Ranking (C2F-FAR) framework for unsupervised long document summarization, and \cite{jain2022improving}, who leveraged the Kullback-Leibler based summarization. Different but general, model-comparative approaches were, for example, undertaken by \cite{rehman2022analysis} and \cite{mahmoud2022novel}.

BERT was, although on other datasets, also used for text summarization by \cite{zhang2019pretraining}, who used the model for the generation of the abstractive summarization output, and \cite{koniaris2023evaluation}, who worked on greek legal texts. Other legal domain automatic text summarization techniques were studied by, among, but not limited to, \cite{liu2019text}, who proposed a general framework for extractive and abstractive models, and \cite{sheik2021deep}, who focused on improvements through preparation mechanisms and throughout various baselines.

The effects of deliberately introducing noise to the specific task of text summarization were studied by \cite{yousefi2017text}, who presented extractive summarization methods based on term-frequency. \cite{liu2020noisy} used noise to better model uncertainty during training with a student and a teacher model interacting with each other. Both reported improvements using noise at the respective steps in the model training processes. Our work then takes a more fundamental and systematic approach, thoroughly studying the different effects of noise on the different steps of the training process.

\section{Methodology}
\label{section:methodology}

In this section, we first describe how we add noise to the language models to regularize them. Afterwards, we briefly touch upon the models we use for our two tasks, \textit{joint named entity recognition and relation extraction} and \textit{text summarization}.

\subsection{Regularization by adding noise}

The parameter matrices of a language model are denoted as $[\bm{W}_1,\bm{W}_2, ..., \bm{W}_N]$, where $N$ is the number of parameter matrix types. As shown in \cite{wu2022noisytune}, the perturbed version $\tilde{\bm{W}}_i$ of the parameter matrix $\bm{W}_i$ is defined as

\begin{equation}
    \tilde{\bm{W}}_i = \bm{W}_i + \text{U}\left(- \frac{\lambda}{2}, \frac{\lambda}{2} \right) \cdot \sigma \left( \bm{W}_i \right),
    \label{eq:noisytune}
\end{equation}


\noindent where $\text{U}\left(a, b \right)$ represents uniformely distributed noise ranging from $a$ to $b$, $\lambda$ is the fine-tunable hyperparameter that controls the noise intensity, and $\sigma \left( c \right)$ denotes the standard deviation of $c$. Clearly, parameters with a larger variance are subject to stronger noise due to the multiplication of the noise term with the standard deviation $\sigma \left( c \right)$.

In addition to simply adding such noise to all parameters, as seen in \cite{wu2022noisytune}, we investigate how the performance of the downstream task is affected when we only partially inject the noise term into certain parts of the model.
More precisely, we add noise to either the bias term, the weight term, or both with different intensities. Furthermore, during the relation extraction downstream task, we add noise to the residual connection and the layer normalization step. We also divide the BERT \cite{devlin2018bert} encoder into two separate \textit{layer zones} to add noise in different intensities, as \cite{tenney-etal-2019-bert} theorized that BERT solves various language understanding tasks at different layer depths. On the other hand, during the text summarization task, we insert noise separately into the encoder and decoder parts of the model.

Therefore, we adjust Equation~\ref{eq:noisytune} by only perturbing a certain part of the parameter matrix $\bm{W}_i$:

\begin{equation}
    \tilde{\bm{W}}^{\text{loc}}_i = \bm{W}^{\text{loc}}_i + \text{U}\left(- \frac{\lambda}{2}, \frac{\lambda}{2} \right) \cdot \sigma \left( \bm{W}^{\text{loc}}_i \right),
    \label{eq:partual_noise}
\end{equation}

\noindent where $\bm{W}^{\text{loc}}_i$ is the localized, i.e. restricted to certain parts of the model, parameter matrix. Then, $\tilde{\bm{W}}^{\text{loc}}_i$ is the perturbed localized parameter matrix. 

\subsection{Joint named entity recognition and relation extraction}

The \textit{joint named entity recognition and relation extraction} task is defined as extracting entities from a text segment, mostly sentences, and linking them together afterwards. Given this sentence from \cite{deusser2022kpiedgar},

\begin{tcolorbox}[notitle,boxrule=0pt,
boxsep=0pt,left=0.6em,right=0.6em,top=0.5em,bottom=0.5em,
colback=gray!10,
colframe=gray!10]
``In 2021 and 2020 the $\underset{\text{{\color{unibonnblue}{kpi}}}}{\text{{\color{unibonnblue}{total net revenue}}}}$ was \$$\underset{\text{{\color{unibonnblue}{cy}}}}{\text{{\color{unibonnblue}{100}}}}$ million and \$$\underset{\text{{\color{unibonnblue}{py}}}}{\text{{\color{unibonnblue}{80}}}}$ million, respectively.''
\end{tcolorbox}

\noindent one should extract and find the relations

\begin{tcolorbox}[notitle,boxrule=0pt,
boxsep=0pt,left=0.6em,right=0.6em,top=0.5em,bottom=0.5em,
colback=gray!10,
colframe=gray!10]
$\underset{\text{{\color{unibonnblue}{kpi}}}}{\text{{\color{unibonnblue}{total net revenue}}}} - \underset{\text{{\color{unibonnblue}{cy}}}}{\text{{\color{unibonnblue}{100}}}}$,
$\underset{\text{{\color{unibonnblue}{kpi}}}}{\text{{\color{unibonnblue}{total net revenue}}}} - \underset{\text{{\color{unibonnblue}{py}}}}{\text{{\color{unibonnblue}{80}}}}$.
\end{tcolorbox}

To solve the joint named entity recognition and relation extraction task, we employ the model introduced in \cite{kpibert} titled KPI-BERT. Said model has three main building blocks: A BERT-based sentence encoder, a named entity recognition decoder, and a relation extraction decoder. 

To give more detail, given an input sentence tokenized using WordPiece \cite{schuster2012japanese}, we utilize the pre-trained BERT model to obtain the encoded token embeddings. Subsequently, we employ a pooling function to generate word representations by combining the embeddings of individual subwords with a trainable recurrent neural network (RNN) pooling mechanism introduced by \cite{kpibert}. The RNN pooling mechanism is built on a bidirectional gated recurrent unit (GRU) \cite{cho2014properties}. Additionally, we employ conditional label masking to sequentially tag entities before classifying their relations. 

\subsection{Text summarization}

The \textit{text summarization} task involves generating a shorter version of a given text while preserving its vital information. It can be approached through extractive or abstractive methods. Extractive methods involve selecting and combining critical sentences from the original text. 

In contrast, abstractive methods involve generating new sentences that capture the essence of the original text, which is the approach we choose to solve. We pick the Longformer-Encoder-Decoder \cite{beltagy2020longformer} setup and pre-trained model to solve the text summarisation task. It is a Longformer variant with encoder and decoder transformer stacks, but the encoder uses Longformer's efficient local and global attention model instead of the initial fully self-attentive one. Additionally, the decoder uses full self-attention for the encoded token and previously decoded locations.

\section{Experiments}
\label{section:experiments}

Here, we first shed some light on the two datasets for our two tasks, namely \textit{joint named entity recognition and relation extraction} and \textit{text summarization}, and how we evaluated these two. Then, we take a closer look at the results and sum them up. All experiments were conducted on a single Nvidia GeForce RTX 2060 SUPER, and the code is implemented in PyTorch.


We execute the experiments within four phases, starting without noise for the downstream tasks. Second, we add noise to all weights as seen in \cite{wu2022noisytune}. These two approaches can be seen as our two baselines. Finally, we arrived at our results by adding noise only to the bias or weight parts. For joint named entity recognition and relation extraction, noise is additionally added to the residual connections and layer normalization and the layer zones. For text summarization, noise is added separately to the encoder and decoder.

Furthermore, to not cherry-pick any particularly well-performing results, as warned about in \cite{trosten2023questionable}, we run each configuration five times with a different seed each and compare the average of these runs.

\subsection{Data and downstream tasks}
\label{subsection:data_and_downstream_tasks}

The joint named entity recognition and relation extraction, defined in the KPI-EDGAR dataset \cite{deusser2022kpiedgar}, aims to extract information from financial documents, including key performance indicators. It holds a total of 1,355 sentences holding 4,522 entities and 3,841 relations, and is split into a training, validation, and test set, encompassing 969, 146, and 240 sentences each. The named entity recognition part is realized in finding numerical and non-numerical entities, whereas the relation extraction part links the found entities to allow for a meaningful extraction of them. \cite{deusser2022kpiedgar} also defined an \textit{adjusted} F$_1$ score capturing when an entity is only partially found, which we will also be using.

For this \textit{adjusted} F$_1$ score, the \textit{true positives} ($\text{tp}$), \textit{false negatives} ($\text{fn}$), and \textit{false positives} ($\text{fp}$) of a relation $r$ between entity $i$ and $j$ are given by:

\begin{align}
    \text{tp}_r &= \frac{1}{2} \left( \frac{o_i}{n_{i, \text{gt}}} + \frac{o_j}{n_{j, \text{gt}}} \right) \\
    \text{fn}_r &= 1 - \text{tp}_r \\
    \text{fp}_r &= \frac{1}{2} \left( \frac{n_{i, \text{pred}} - o_i}{n_{i, \text{pred}}} +  \frac{n_{j, \text{pred}} - o_j}{n_{j, \text{pred}}} \right), \label{eqf1}
\end{align}

\noindent where 

\begin{align}
    o_i := \left| e_{i, \text{pred}} \cap e_{i, \text{gt}} \right| , \label{eqoverlap}
\end{align}

\noindent is the overlap/intersection $o_i$ of an entity prediction $i$ and its ground truth and $e_{i, \text{pred}}$ and $e_{i, \text{gt}}$ is the set of all token identifiers for the entity prediction and ground truth, respectively. The operation $|\cdot|$ calculates the size of a given set.

For the text summarization task, we consider the BillSum dataset \cite{kornilova-eidelman-2019-billsum}, which aggregates U.S. congressional and California state bills and was split into 18,949 
training bills and 3,269 test bills. We evaluate our results using the ROUGE metrics proposed in \cite{lin-2004-rouge}. Furthermore, to make a convenient comparison possible, we average the values of the ROUGE--1 F$_1$, ROUGE--2 F$_1$, ROUGE--L F$_1$ and ROUGE--L--sum and title it \textit{ROUGE--Average}.

\subsection{Results}

\begin{table}[t]
\centering
    \rowcolors{2}{gray!10}{white}
\begin{tabular}{lcc}
\toprule
\rowcolor{white} Noise added to & $\lambda$ & F$_1$ in  \%
\\
\midrule
Nothing &0 & 40.696
\\

All & 0.81 & 42.824
\\

\textbf{Bias} & \textbf{0.41} & \textbf{43.672}
\\

Weights & 0.50 & 43.084
\\

Add$\&$Norm & 0.2 & 43.411
\\

Layer zones & 0.9 & 42.200
\\
\bottomrule

\end{tabular}


    \caption{Results of adding noise to KPI-BERT \cite{kpibert}, evaluated on KPI-EDGAR \cite{deusser2022kpiedgar}.  Adding noise to all parameters is equivalent to the approach from \cite{wu2022noisytune}. Add\&Norm 
    refers to the process of adding noise to residual connections and layer normalization.}
    \label{tab:kpiedgar_main}
\end{table}

\begin{table}[t]
\centering
    \rowcolors{2}{gray!10}{white}
\begin{tabular}{lcc}
\toprule
\rowcolor{white} Noise added to & $\lambda$ & ROUGE--Average in  \%
\\
\midrule
Nothing &0 &  34.148
\\

All & 0.3 & 35.131
\\

Bias & 0.4 &  34.854
\\

Weights & 0.1 &  34.703
\\

Encoder & 0.1 &  34.709
\\

\textbf{Decoder} & \textbf{0.8} & \textbf{35.534}
\\
\bottomrule

\end{tabular}
    \caption{Results of adding noise to the Longformer-Encoder-Decoder \cite{beltagy2020longformer}, evaluated on BillSum \cite{kornilova-eidelman-2019-billsum}.  Adding noise to all parameters is equivalent to the approach from \cite{wu2022noisytune}.}
    \label{tab:billsum_main}
\end{table}

As seen in Table~\ref{tab:kpiedgar_main} and \ref{tab:billsum_main}, adding noise, in general, does help the models generalize better. Thus, we can confirm the findings in \cite{wu2022noisytune}. Additionally, we are able to improve upon that by exposing only certain parts of the models to noise.

Interestingly, there is no simple \say{best approach} on where to add noise in a model, as seen in the different best-performing noise locations of Table~\ref{tab:kpiedgar_main} and \ref{tab:billsum_main}. Therefore, this is another hyperparameter that has to be fine-tuned. In \cite{wu2022noisytune}, they only tested a limited range, i.e. $\lambda \in \{0, 0.05, 0.1, 0.15, 0.2, 0.25, 0.3\}$, and they found that $0.1 \leq \lambda \leq 0.15$ were optimal. Contrasting this, in our experiments, the best results are obtained with noise intensities significantly larger than this, as demonstrated in Table~\ref{tab:kpiedgar_main} and \ref{tab:billsum_main}.

Still, we found that our approach boosted both performances. In the case of KPI-BERT \cite{kpibert} on KPI-EDGAR \cite{deusser2022kpiedgar}, we achieve a remarkable increase of \textbf{2.977}\% F$_1$ compared to the non-perturbed model and an increase of \textbf{0.849}\% F$_1$ compared to the NoisyTune approach \cite{wu2022noisytune}. On BillSum \cite{kornilova-eidelman-2019-billsum}, our approach improved the performance of the Longformer-Encoder-Decoder \cite{beltagy2020longformer} by \textbf{1.387}\% and \textbf{0.403}\% compared to the non-perturbed model and the NoisyTune baseline, respectively.

\section{Conclusion and Future Work}
\label{section:conclusion}

In this paper, we study the effect of controlled randomness on transformer models and how such noise, introduced into various parts of the model as shown in Equation~\ref{eq:partual_noise}, can be seen as a potent regularization tool for ever-increasing language models. We studied two different downstream tasks, namely \textit{joint named entity recognition and relation extraction} and \textit{text summarization}, and found that if certain parts of their respective transformer models are infused with a noise component, we can increase their performance significantly. In the first task, three out of the four scenarios, i.e. adding noise to either the bias term, the weights, or to residual connections and layer normalization, were successful and demonstrated additional enhancements over our already string baseline introduced in \cite{wu2022noisytune}. In the second task, we can still improve on this baseline, but only one of our scenarios offers an improvement, namely adding noise the whole decoder and leaving the encoder as it is.

There are two further conclusions from our results. One is that the optimal value of the hyperparameter $\lambda$, as defined in Equation~\eqref{eq:partual_noise}, depends heavily on the dataset and on the task. The other is that in the paper introducing our baseline NoisyTune \cite{wu2022noisytune}, they only tested the range $\lambda \in \{0, 0.05, 0.1, 0.15, 0.2, 0.25, 0.3\}$, which should have been extended further to achieve even better results, as seen in Table~\ref{tab:kpiedgar_main} and \ref{tab:billsum_main}.

In future work, we plan to tackle even more downstream tasks, like named entity recognition, natural language inference, or sentiment analysis, to see if we can achieve such positive results on these natural language processing applications as well. One could also apply this regularization technique to any pre-trained large language model like Llama \cite{touvron2023llama} or Bloom \cite{workshop2023bloom} and experiment on how these can then handle themselves. It might partly alleviate the flaw of requiring large datasets for fine-tuning  when training data is sparse. 

Another interesting idea would be trying out different noise distributions than the uniformly distributed one seen in Equation~\eqref{eq:partual_noise}. One candidate for this could be a Gaussian distribution with fat tails, as this does not have such extreme cut-off points as a uniform distribution.

Furthermore, it would be immensely interesting to see if some low-resource languages can benefit from such a regularization approach. We theorize that in this case, one might see even larger increases in performance, as the \say{seen language} imbalance of multilingual models, i.e. the distribution of different languages in the training dataset, is usually quite pronounced and heavily skewed towards English, as seen in e.g. \cite{conneau-etal-2020-unsupervised} or \cite{laurenccon2022bigscience}.

\section*{Acknowledgments}

This research has been funded by the Federal Ministry of Education and Research of Germany and the state of North-Rhine Westphalia as part of the Lamarr-Institute for Machine Learning and Artificial Intelligence. 

\bibliographystyle{plainnat}
\bibliography{references}

\end{document}